\documentclass[lettersize,journal]{IEEEtran}

\usepackage[caption=false,font=normalsize,labelfont=sf,textfont=sf]{subfig}
\usepackage{graphicx}
\usepackage{standalone}
\usepackage{tikz}
\usetikzlibrary{shapes,backgrounds,shapes.geometric}
\usepackage{circuitikz}
\tikzset{font={\fontsize{9pt}{9}\selectfont}}
\usepackage{pgfplots}
\usepgfplotslibrary{fillbetween,colormaps,groupplots}
\pgfplotsset{compat=newest}

\usepackage{tabularx}
\usepackage{makecell}
\usepackage{booktabs}
\newcolumntype{Y}{>{\centering\arraybackslash}X}
\newcolumntype{s}{>{\hsize=0.7\hsize}Y}

\usepackage{siunitx}
\usepackage{amsmath}
\usepackage{amsfonts}
\usepackage{scalerel}
\usepackage{amssymb}
\usepackage{bm}

\usepackage{algorithm}
\usepackage{algorithmicx}

\DeclareMathOperator{\sign}{\text{sgn}}

\usepackage{cite}
\usepackage{jabbrv}

\usepackage[utf8]{inputenc}
\usepackage{enumitem}

\date{\today}

\usepackage{hyperref}
\hypersetup{
	pdfauthor={Julian Schumann},
	pdfcreator={Julian Schumann},
	pdfproducer={},
	pdfsubject={First project for PhD},
    colorlinks={true},
    linkcolor={blue},
    citecolor={blue},
    urlcolor={red!50!white},
    hyperindex={true}
}
\makeatletter
\hypersetup{pdftitle=\@title}
\makeatother
\usetikzlibrary{external}
\title{Benchmarking Behavior Prediction Models \\ in Gap Acceptance Scenarios}
\author{Julian F. Schumann, Jens Kober, Arkady Zgonnikov
\thanks{Manuscript received October 27, 2022; revised manuscript received January 02, 2023; Accepted Februar 06, 2023; \copyright 2023 IEEE. Personal use of this material is permitted.  Permission from IEEE must be obtained for all other uses, in any current or future media, including reprinting/republishing this material for advertising or promotional purposes, creating new collective works, for resale or redistribution to servers or lists, or reuse of any copyrighted component of this work in other works.}
\thanks{The authors are with the Department of Cognitive Robotics, Delft University of Technology, Delft, Zuid Holland 2628 CD, The Netherlands (e-mail: j.f.schumann@tudelft.nl; j.kober@tudelft.nl; a.zgonnikov@tudelft.nl) \textit{(Corresponding Author: Julian Schumann)}}
\thanks{The source code, trained models, and data can be found online at  \href{https://github.com/julianschumann/Framework-for-benchmarking-gap-acceptance}{public Github repository}}}

\begin{document}
\twocolumn
\maketitle

\begin{abstract}
    Autonomous vehicles currently suffer from a time-inefficient driving style caused by uncertainty about human behavior in traffic interactions. 
    Accurate and reliable prediction models enabling more efficient trajectory planning could make autonomous vehicles more assertive in such interactions. However, the evaluation of such models is commonly oversimplistic, ignoring the asymmetric importance of prediction errors and the heterogeneity of the datasets used for testing. We examine the potential of recasting interactions between vehicles as gap acceptance scenarios and evaluating models in this structured environment. To that end, we develop a framework aiming to facilitate the evaluation of any model, by any metric, and in any scenario. We then apply this framework to state-of-the-art prediction models, which all show themselves to be unreliable in the most safety-critical situations.
\end{abstract}
\section{Introduction}
\IEEEPARstart{S}{uccessfully} implementing autonomous driving is one of the key technical challenges faced by the automotive industry as well as large parts of the research community, with tens of billions of dollars invested in recent years towards this goal~\cite{holland-letz_mobilitys_2021}. The provision of those funds is motivated by several benefits promised by this technology. The foremost of these is safer driving, expressed by a significant decrease in accidents and, correspondingly, a reduction of bodily harm and financial losses. Additional advantages are also expected, such as more accessible mobility for people unable to drive or an easing of road congestion and traffic~\cite{brar_impact_2017, meyer_autonomous_2017, pisarov_future_2021}.

\begin{figure}[!ht]
\centering
\includegraphics[]{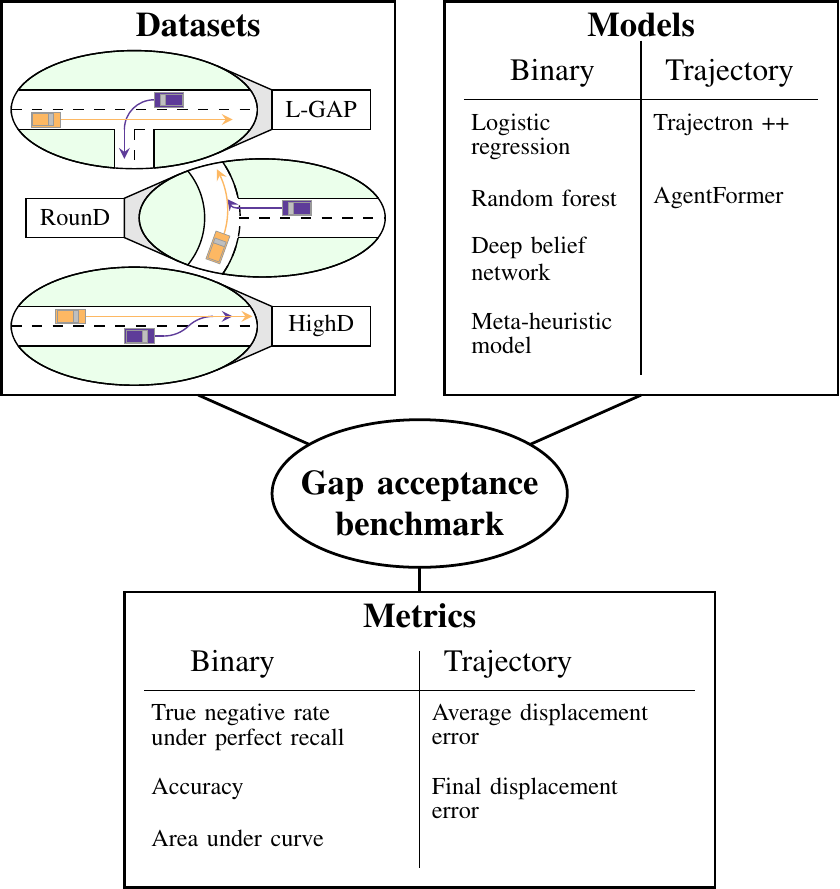}
\caption{The proposed framework allows researchers to evaluate the performance of various prediction models for human behavior according to several metrics on different datasets that include gap acceptance scenarios. In this work, three datasets, six models, and four metrics (dealing both with binary and trajectory predictions) are implemented.}
\label{fig:overview}
\end{figure}

But despite all these investments, autonomous vehicles still suffer from many problems preventing widespread use~\cite{milford_self-driving_2020, wang_safety_2020}. One such problem is their timidity in interactions with human traffic participants, caused by the uncertainty about the future behavior of those human agents. This uncertainty can prevent the autonomous vehicle from taking the most time-efficient actions if the resulting probability of a crash or near-crash is too high, resulting in the cautious driving style observed. Paradoxically, this can also be a safety risk, as such caution by an autonomous vehicle is often not expected by the surrounding humans, which can result in accidents such as being rear-ended~\cite{milford_self-driving_2020, sinha_crash_2021}.

To reduce this uncertainty and to allow for a more efficient driving style without compromising on safety requirements, behavior prediction models can be used~\cite{sadigh_planning_2016, mozaffari_deep_2022}, which project the future position of traffic participants. Those can range from models able to deal with any kind of traffic participant~\cite{salzmann_trajectron_2020, giuliari_transformer_2021, yuan_agentformer_2021} to others focused on predicting the behavior of a specific kind of participant, such as cars~\cite{diehl_graph_2019,kolekar_human-like_2020, huang_hyper_2022, cui_lookout_2021,chandra_using_2022,sun_complementing_2021, cao_leveraging_2022} or pedestrians~\cite{bighashdel_towards_2020,sadeghian_sophie_2019, kothari_interpretable_2021,camara_pedestrian_2021, yue_human_2022}.

However, the utility of those models---primarily designed to minimize the necessary trade-off between safety and efficiency in trajectory planning---is questionable, as the common methods for their evaluation diverge from the models' purpose. First, most common metrics for evaluating prediction models, such as the final or average displacement error, ignore that the consequences of a false prediction are inherently asymmetric\cite{ivanovic_injecting_2022, farid_task-relevant_2022}. For example, on a highway, wrong longitudinal predictions are far less dangerous than wrong lateral predictions, which might result in an autonomous vehicle reacting to a lane change too late. Similarly, such metrics also lack the ability to evaluate how good models are at capturing distinct human behaviors~\cite{srinivasan_comparing_2021, srinivasan_beyond_2022}.
Second, the common approach of randomly selecting test cases from datasets~\cite{salzmann_trajectron_2020, yuan_agentformer_2021} is problematic due to the heterogeneity of those datasets, which typically include samples that can vary widely in their importance and difficulty. Such samples can range from a single vehicle following a lane to complex space-sharing conflicts with multiple agents at unsignalized intersections, where the behavior of human agents is often multi-modal and can change rapidly. Rare edge cases, where some traffic participants are very aggressive or even violate traffic rules and accidents are far more likely~\cite{noh_probabilistic_2018, zhang_adversarial_2022}, are also possible. But with randomly selected test cases, potentially poor performance in the most important situations can be compensated by good performance in less important but more numerous ones.
For these reasons, a model that is unreliable in actual safety-critical situation might still appear promising, making the whole evaluation meaningless and hampering further progress.

One possible approach to overcome these issues is including a path planning algorithm in evaluations, as suggested by Ivanovic and Pavone~\cite{ivanovic_injecting_2022}. However, this adds further computational loads to an evaluation and only addresses the symmetry of common metrics, neglecting the varying difficulty and importance between testing samples. To cover both these problems, we suggest instead narrowing the evaluation to the most critical situations. In particular, we focus the evaluation of behavior prediction models on \emph{gap acceptance} scenarios, a concept that encompasses most of the safety critical interactions between autonomous vehicles and humans~\cite{markkula_defining_2020}. In a gap acceptance scenario, an autonomous vehicle follows a particular trajectory over which a second traffic participant (e.g., a pedestrian or another vehicle) can move either in front of or behind the autonomous vehicle. Here, the first option (i.e., the human accepting the gap) would require the autonomous vehicle to potentially alter its trajectory planning, while the latter one of rejecting the gap would not. Due to the narrow focus, estimating the importance and difficulty of a particular situation can become much more straightforward. Additionally, as the human has only two options to decide between, such gap acceptance scenarios allow the usage of simple binary prediction models to estimate if the human behavior requires an adjustment of trajectory planning.

Many binary prediction models have been developed for gap acceptance scenarios. However, those mostly focus and are trained on a specific scenario, such as the street crossing behavior of pedestrians~\cite{jayaraman_multimodal_2021, pekkanen_variable-drift_2022, theofilatos_cross_2021, yang_predicting_2022}, the crossing behavior of cars at intersections~\cite{mafi_analysis_2018, abhishek_generalized_2019, zgonnikov_should_2022, arafat_stop_2021, hu_causal-based_2022}, or lane change decisions on high ways~\cite{balal_comparative_2018,xie_data-driven_2019,das_nonparametric_2020, khelfa_predicting_2023, mozaffari_early_2022}. Additionally, the development of those models still suffers from similar problems as the trajectory prediction models, such as the anisotropy of common metrics like accuracy. Likewise, a random selection of test cases~\cite{zgonnikov_should_2022, pekkanen_variable-drift_2022} and neglect of the varying importance of different samples are also common. Additionally, in contrast to trajectory prediction models, which are commonly compared to each other on accepted benchmarks (such as on the ETH dataset~\cite{salzmann_trajectron_2020, giuliari_transformer_2021, yuan_agentformer_2021} when predicting pedestrian crowds), an equivalent benchmark does not exist for binary prediction models~\cite{rudenko_human_2020}. Instead, those models are mostly trained and tested on datasets exclusive to the respective work and are---if at all---only compared against a small number of other selected models~\cite{kadali_models_2015, jayaraman_multimodal_2021, pekkanen_variable-drift_2022, zgonnikov_should_2022, hu_causal-based_2022, arafat_stop_2021, yao_deep_2021, nagalla_analyzing_2017, yang_examining_2019, pool_crafted_2021}.

Our goal in this work is to overcome these limitations of the current literature on both binary and trajectory prediction models and enable a meaningful evaluation of these models in gap acceptance scenarios. Such an evaluation cannot only make the development of trajectory prediction models more goal-oriented but also help determine to what extent the inclusion of specialized binary prediction models can improve the performance and reliability of general trajectory prediction models. To that end, this paper makes three main contributions:
\begin{itemize}
    \item We develop a formal description of the gap acceptance process that applies to all possible gap acceptance scenarios. This description includes a detailed timeline of gap acceptance (Section~\ref{sec:characteristic_time_points}), which serves as a foundation for methods to estimate the criticality concerning the safety of each sample, which is a fundamental requirement for selecting meaningful test cases.
    \item We devise a framework for evaluating behavior prediction models in gap acceptance scenarios. This novel framework allows the integration of varied gap acceptance datasets, models, and evaluation metrics (Section~\ref{sec:framework}). It is inspired by similar works by M{\"u}ller et al. for computer-based image retrieval algorithms~\cite{muller_framework_2003}, by Zaffar et al. in the field of visual place recognition~\cite{zaffar_vpr-bench_2021}, or Cao et al. for evaluating robustness to adversarial attacks of trajectory prediction models~\cite{cao_advdo_2022}. This approach would allow one to test and compare models in a chosen environment more easily, no matter if they make binary or trajectory predictions. For example, the proposed framework enables evaluating trajectory prediction models using binary metrics in gap acceptance scenarios, something that has not been previously investigated. Simultaneously, the framework allows precise control over the splitting of the data into training and testing samples to comprehensively evaluate the models' reliability in the most difficult gap acceptance situations.
    \item Using the proposed framework, we compare six prediction models in their performance on three gap acceptance datasets (Section~\ref{sec:Used_modules} and Figure~\ref{fig:overview}). First, this demonstrates the general viability of our proposed method, by including models making and metrics evaluating both binary and trajectory predictions. Second, it allows us to test the reliability of those models specifically in safety-critical edge cases, which is currently missing from the literature. Third, it allows the testing of the hypothesis that including dedicated binary models for gap acceptance problems could improve the performance of state-of-the-art trajectory prediction models.
    Lastly, this gives researchers easy access to an already implemented baseline to compare their own behavior prediction models to. 
\end{itemize}

\section{Defining gap acceptance}
\label{sec:characteristic_time_points}
\begin{figure*}[!ht]
\centering
\includegraphics[]{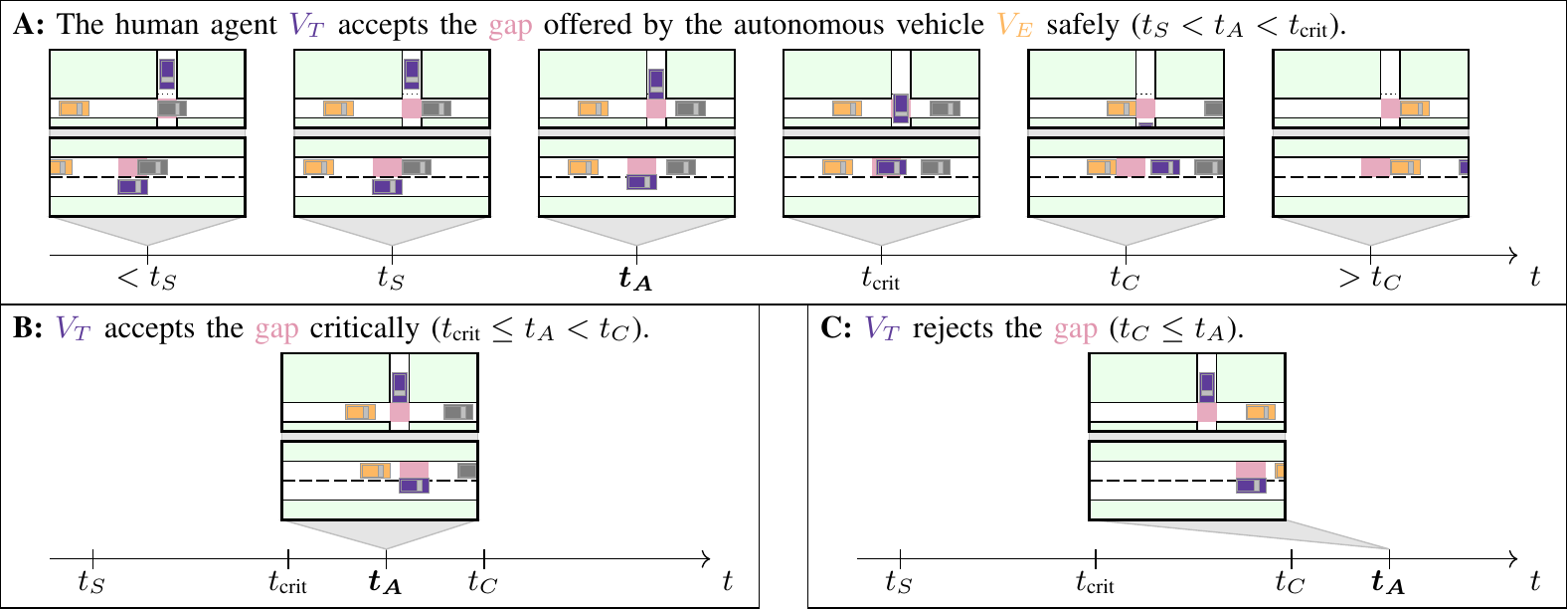}
\caption{
The characteristic time-points of the gap acceptance process---defined by the relation of the agents to the contested space (in purple)---of two different examples of gap acceptance, intersection crossing (upper panels) and lane changing (lower panels). In both examples, the autonomous vehicle $V_E$ (in red) offers a gap to the human-driven vehicle $V_T$ (in blue). In total, three cases are possible (\textbf{A} -- \textbf{C}), depending on $t_A$, i.e., the time the target vehicle enters the contested space. In \textbf{B}, the accepted gap decision by the human is considered to be unsafe, as $V_E$ cannot guarantee the avoidance of a crash, having potentially not enough time for braking. Meanwhile, in \textbf{C}, it might be possible that $V_T$ crashes into $V_E$.}
\label{fig:timeline}
\end{figure*}

To estimate the difficulty and control the importance of prediction tasks over disparate datasets, a coherent formal definition of gap acceptance scenarios is needed. Here we propose such a definition. 

In a gap acceptance scenario, an autonomous vehicle $V_E$---also referred to as the ego-vehicle---plans to follow along a certain trajectory $P_E$ along which it has the right of way. This trajectory overlaps with the trajectory $P_T$ of another, human-controlled vehicle $V_T$ (also named target vehicle). Such an overlap might, for example, happen at unsignalized intersections, where the agents move along crossing streets or on highways, where $V_T$ wants to merge into the faster lane along which $V_E$ is driving. In such situations, $V_T$ can decide to move onto $P_E$ either in front of or behind $V_E$, i.e., to accept or reject the gap offered by $V_E$.
We assume that $V_E$ has the right of way along $P_E$, as otherwise, traffic rules would obligate it to preemptively yield.  

Under these conditions, a gap acceptance scenario is characterized by the spatiotemporal relation between the agents towards the so-called contested space~\cite{markkula_defining_2020}. There, the trajectories $P_E$ and $P_T$ would start to overlap, making this the location of a potential collision. An example is the overlap of two crossing lanes at an intersection. However, in specific scenarios (such as changing lanes on highways), the exact location of the meeting point of $P_E$ and $P_T$ can be at the discretion of the human agent $V_T$ and therefore be unknown before the actual decision. In such cases, we then place the contested space under the assumption that $V_T$ would decide to accept the gap immediately. For example, in the scenario of highway lane changes, the contested space would therefore move in parallel to $V_T$, only stopping to move once $V_T$ starts to enter the lane of $V_E$. 

The following time points then characterize the gap acceptance process (illustrated together with the contested space in Figure~\ref{fig:timeline}):
\begin{itemize}[align = left, leftmargin=1cm, itemindent = 0cm, labelwidth = 0.9cm, labelsep = 0.1cm]
    \item[$t_S$:] At the \textbf{s}tarting time  $t_S$, there is no longer any other vehicle along $P_E$ in between $V_E$ and the contested space. This is primarily the case when the vehicle preceding $V_E$ leaves the contested space, but other options are imaginable, like the vehicle in front of $V_E$ leaving $P_E$.
    \item[$t_C$:]At $t_C$, $V_E$ starts to enter the contested space, \textbf{c}losing the gap.
    \item[$t_{\underline{C}}(t)$:] A prediction of $t_C$ by the ego vehicle, made at $t$, needed to allow gap size estimations during online applications. While this is scenario-dependent, the following condition has to be satisfied so that an open gap can still be characterized as such, even if $V_E$ is moving away from the contested space:
    \begin{equation*}
        \sign \left( t_{\underline{C}}(t) - t \right) = \sign \left( t_C - t \right)
        \label{eq:tchat_cond}\,.
    \end{equation*}
    \item[$t_{\text{crit}}$:] The last time $V_E$ can safely prevent a collision even in the case of malicious behavior by $V_T$; e.g., at this point, a safe braking process could bring $V_E$ to a stop before the intersection. $t_{\text{crit}}$ can be formalized in the following condition:
    \begin{equation}
        \Delta t_D (t) = t_{\underline{C}}(t) - t - t_{\text{brake}}(t) = 0 \label{eq:crit_cond1}
    \end{equation} 
    Here, the required braking time $t_{\text{brake}}$ is not based on the maximum deceleration $V_E$ is technically capable of, but instead, one that is considered safe. The time point $t_{\text{crit}}$ is also the last time a prediction can be considered useful for further trajectory planning.
    \item[$t_A$:] At $t_A$, $V_T$ enters the contested space, potentially \textbf{a}ccepting the gap.
\end{itemize}

We count $V_T$ as rejecting the gap if $V_E$ is allowed to move first onto the contested space, i.e., if $t_C \leq t_A$. If this is not the case and the human moves first ($t_A < t_C$), the gap is considered accepted.

\section{Framework for benchmarking gap acceptance models}
\label{sec:framework}
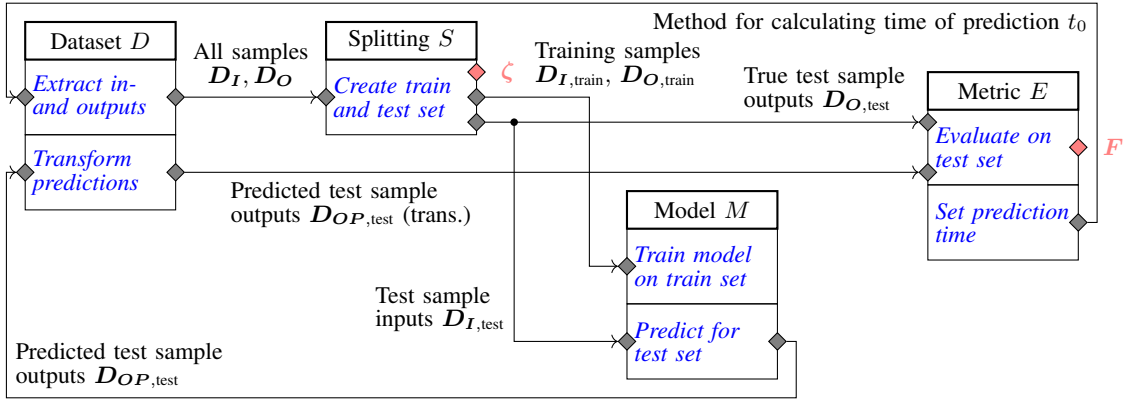
\begin{figure*}[!ht]
\centering
\tikzset{external/export next=false}
\tikzsetnextfilename{Figure_1}
\begin{tikzpicture}

    \begin{scope}[xshift = 1cm, yshift = - 2cm]
    \draw[black, thick] (0,-0.5) rectangle(2,-1) node[pos=0.5, align=center, text width = 1.8cm]
    {Dataset $D$};
    \draw[black] (0,-1) rectangle(2,-2) node[pos=0.5, align=left, text width = 1.8cm] (DI)
    {\textit{\hyperref[sec:DI]{Extract in- \\ and outputs}}};
    \node [draw = black, scale=0.5 ,diamond, fill=black!50] (DI_out) at (2, -1.5) {};
    \node [draw = black, scale=0.5 ,diamond, fill=black!50] (DI_in) at (0, -1.5) {};
    \draw[black] (0,-2) rectangle(2,-3) node[pos=0.5, align=left, text width = 1.8cm]
    {\textit{\hyperref[sec:DT]{Transform\\predictions}}};
    \node [draw = black, scale=0.5 ,diamond, fill=black!50] (DT_in) at (0, -2.5) {};
    \node [draw = black, scale=0.5 ,diamond, fill=black!50] (DT_out) at (2, -2.5) {};
    \end{scope}
    
    \begin{scope}[xshift = 5cm, yshift = - 3cm]
    \draw[black, thick] (0,0.5) rectangle(2,0) node[pos=0.5, align=center, text width = 1.8cm]
    {Splitting $S$};
    \draw[black] (0,0) rectangle(2,-1) node[pos=0.5, align=left, text width = 1.8cm]
    {\textit{\hyperref[sec:SM]{Create train\\ and test set}}};
    \node [draw = black, scale=0.5 ,diamond, fill=black!50] (S_in) at (0, -0.5) {};
    \node [draw = black, scale=0.5 ,diamond, fill=black!50] (S_out1) at (2, -0.5) {};
    \node [draw = black, scale=0.5 ,diamond, fill=black!50] (S_out2) at (2, -0.833) {};
    \node [draw = black, scale=0.5 ,diamond, fill=red!50!white] (S_outS) at (2, -0.166) {};
    \node[right, align= left, red!50!white] at (2.2, -0.166)
    {$\bm{\zeta}$};
    \end{scope}

    \begin{scope}[xshift = 9cm, yshift = - 4.25cm]
    \draw[black, thick] (0,-0.5) rectangle(2,-1) node[pos=0.5, align=center, text width = 1.8cm]
    {Model $M$};
    \draw[black] (0,-1) rectangle(2,-2) node[pos=0.5, align=left, text width = 1.8cm]
    {\textit{\hyperref[sec:MT]{Train model \\ on train set}}};
    \node [draw = black, scale=0.5 ,diamond, fill=black!50] (MT_in) at (0, -1.5) {};
    \draw[black] (0,-2) rectangle(2,-3) node[pos=0.5, align=left, text width = 1.8cm]
    {\textit{\hyperref[sec:MP]{Predict for \\ test set}}};
    \node [draw = black, scale=0.5 ,diamond, fill=black!50] (MP_in) at (0, -2.5) {};
    \node [draw = black, scale=0.5 ,diamond, fill=black!50] (MP_out) at (2, -2.5) {};
    \end{scope}
    
    \begin{scope}[xshift = 13cm, yshift = - 3.667cm]
    \draw[black, thick] (0,0.5) rectangle(2,0) node[pos=0.5, align=center, text width = 1.8cm]
    {Metric $E$};
    \draw[black] (0,-1) rectangle(2,-2) node[pos=0.5, align=left, text width = 1.8cm]
    {\textit{\hyperref[sec:ET]{Set prediction time}}};
    \node [draw = black, scale=0.5 ,diamond, fill=black!50] (ET_out) at (2, -1.5) {};
    \draw[black] (0,0) rectangle(2,-1) node[pos=0.5, align=left, text width = 1.8cm]
    {\textit{\hyperref[sec:EV]{Evaluate on \\ test set}}};
    \node [draw = black, scale=0.5 ,diamond, fill=black!50] (EV_in1) at (0, -0.166) {};
    \node [draw = black, scale=0.5 ,diamond, fill=black!50] (EV_in2) at (0, -0.833) {};
    \node [draw = black, scale=0.5 ,diamond, fill=red!50!white] (EV_out) at (2, -0.5) {};
    \node[right, align= left, red!50!white] at (2.2, -0.5)
    {$\bm{F}$};
    \end{scope}
    
    \draw [->] (DI_out) -- (S_in);
    \node[align= center, above] at (4, -3.5)
    {All samples \\ $\bm{D_I}, \bm{D_O}$};
    
    \draw [->] (S_out1) -- (8.5, -3.5) --  (8.5, -5.75)  -- (MT_in); 
    \node[above right, align= left] at (7.65, -3.5)
    {Training samples \\ $\bm{D}_{\bm{I},\text{train}},$ $\bm{D}_{\bm{O},\text{train}}$};
    
    \draw [->] (S_out2) -- (EV_in1);
    \node[above, align= left] at (11.625, -3.833)
    {True test sample \\ outputs  $\bm{D}_{\bm{O},\text{test}}$};
    
    \draw [->] (7.5, -3.833) --  (7.5, -6.75) -- (MP_in); 
    \fill [black] (7.5, -3.833) circle(0.05); 
    \node[align= left, above left] at (7.5, -6.75)
    {Test sample \\ inputs $\bm{D}_{\bm{I},\text{test}}$};
    
    \draw [->] (MP_out) -- (11.25,-6.75) -- (11.25,-7.5)-- (0.75,-7.5) -- (0.75,-4.5) -- (DT_in); 
    \node[align= left, above right] at (0.75, -7.5)
    {Predicted test sample \\ outputs $\bm{D}_{\bm{OP},\text{test}}$};;
    
    \draw [->] (DT_out) -- (EV_in2); 
    \node[align= left, below] at (5.33, -4.5)
    {Predicted test sample \\ outputs $\bm{D}_{\bm{OP},\text{test}}$ (trans.)};
    
    \draw [->] (ET_out) -- (15.25, - 5.167) -- (15.25, -2.25) -- (0.75, -2.25) -- (0.75, - 3.5) -- (DI_in); 
    \node[align= left, below left] at (15.25, -2.25)
    {Method for calculating time of prediction $t_0$};

\end{tikzpicture}
\caption{Functionalities of the proposed framework. To evaluate a model $M$ on a dataset $D$ with the metric $E$ and splitting method $S$, a method for determining the prediction time $t_0$ has to be chosen first (see Section \ref{sec:ET} for a detailed description). This method is used to extract input and output trajectories ($\bm{D}_I$ and $\bm{D}_O$ respectively) from the dataset $D$ (\ref{sec:DI}). Those samples are split in training and testing set using the splitting method $S$ (\ref{sec:SM}), with the training one ($\bm{D}_{I,\text{train}}$, $\bm{D}_{O,\text{train}}$) used to train a model $M$ (\ref{sec:MT}). Subsequently, predictions $\bm{D}_{OP,\text{test}}$ are made for the test samples $\bm{D}_{I,\text{test}}$ with the trained model (\ref{sec:MP}). It might be necessary to transform these predictions into another form (\ref{sec:DT}), before the metric $E$ compares them to the true outputs $\bm{D}_{O,\text{test}}$ (\ref{sec:EV}). These steps produce two outcomes (red diamonds): the similarity $\bm{\zeta}$ between training and test data, provided by $S$, and the model performance $\bm{F}$ according to metric $E$.}
\label{fig:diagram}
\end{figure*}

After defining the fundamental characteristics of a gap acceptance scenario, we will use this groundwork to build a framework for benchmarking gap acceptance models.
This framework should allow for the performance assessment of a prediction model $M$ on a dataset $D$ according to evaluation metric $E$. The following requirements need to be met for such an assessment to be both meaningful and possible for as many of the aforementioned modules as possible:
\begin{enumerate}[label=\textbf{R \arabic*}, align=left, leftmargin=1cm, itemindent = 0cm, labelwidth = 0.9cm, labelsep = 0.1cm]
    \item The time point $t_0$ of a prediction must be controllable, as it influences not only the difficulty of the prediction but also its importance due to changing consequences of a false prediction. \label{req:1}
    \item To evaluate models in critical situations, the framework should allow control over splitting all available samples into training and testing sets. \label{req:2}
    \item Models producing (as well as metrics evaluating) for example binary or trajectory predictions should fit into the framework. Therefore, the framework should allow transformations between those forms of model output.\label{req:3}
\end{enumerate}
Considering these requirements, seven functionalities will constitute the proposed framework. Figure~\ref{fig:diagram} illustrates these functionalities and their mutual dependencies. These functionalities are grouped in four modules (dataset $D$, splitting method $S$, model $M$, and Metric $E$); in this section we describe them in the order in which they are employed in the process of a single evaluation.

\subsection{Setting the prediction time --- Metric $E$} \label{sec:ET}
To satisfy requirement \hyperref[req:1]{\textbf{R 1}}, this functionality enables the selection of the time-point $t_0$ at which the prediction has to be made. As the prediction time influences both the importance of such predictions and the meaningfulness of different metrics (Appendix~\ref{App:details_EV}), this functionality is attached to the metric module.

Currently, three methods are implemented into the framework to determine $t_0$:
\begin{itemize}
    \label{sec:t0_methods}
    \item \emph{Prediction at the initial opening of the gap}: $t_0 = t_S$. The prediction is made when the gap first appears, and this is the baseline most commonly used in the literature~\cite{zgonnikov_should_2022, kadali_models_2015,xie_data-driven_2019}.
    \item \emph{Prediction at gaps with fixed size}: $t_0 = \min \{t \,\vert \, t_{\underline{C}}(t) - t = \Delta t\}$. The prediction is made when the gap offered has a uniform duration $\Delta t$, which should make every prediction equally difficult due to a similar prediction horizon.
    \item \emph{Last useful prediction for critical gaps}: $t_0 = t_{\text{crit}} - t_{\epsilon}$. The prediction is made at the last point in time when it would still be useful, with $t_{\epsilon}$ being used to allow time for calculations.
\end{itemize}
Here, $t_0$ has to be calculated without hindsight knowledge for online predictions at a time when $t_C$ or $t_A$ are not known.
A discussion on the impact of the different approaches on the resulting datasets can be found in Appendix~\ref{App:details_ET}.

\subsection{Extracting input and output --- Dataset $D$} \label{sec:DI}
Next, the input and output data for each sample are extracted from a given trajectory $\bm{X_T}$, which includes positions at different time points $\bm{T}$ from different actors $\bm{V} = \{V_E, V_T, V_1, \hdots \}$:
\begin{equation*}\begin{aligned}
    \bm{X_T} &= \left\{\bm{x} (t) \, \vert \, t \in \bm{T}   \right\},\; \text{where} \; 
    I_{\bm{T}} = \left[ \max \bm{T} , \min \bm{T}  \right]\\
    \bm{x}(t) &= \left\{ \bm{x}_{i}(t) \,\vert\, V_i \in \bm{V}   \right\} \\
    \bm{x}_{i}(t) &= \left(x_i(t), y_i(t)\right) \in \mathbb{R}^2 \label{eq:data_format}
\end{aligned}\end{equation*}
This functionality, requiring access to the raw data from the scenario and thus being part of the dataset module, consists of ten consecutive steps:
\begin{enumerate}
    \item $t_{\underline{C}}$ and $t_{\text{brake}}$ are estimated at every time point in $\bm{T}$, which is scenario-dependent.
    \item The characteristic time points $t_S$, $t_A$ and $t_C$ are determined.  Using the scenario-specific conditions $C_S$, $C_C$, and $C_A$ respectively ($C_S$ is true if the gap is offered, $C_C$ and $C_A$ are true if respectively the ego vehicle $V_E$ and target vehicle $V_T$ are inside the contested space), one then has to to find the specific times $\bm{T}_{C_i} = \left\{ t \,\vert\, C_i(t) \, \forall \, t\in I_{\bm{T}}\right\}$ at which those conditions are true (e.g., $\bm{T}_{C_A}$ is the time during which $V_T$ is inside the contested space); from this, the characteristic time points are extracted:
    \begin{equation}\begin{aligned}
        t_S & = T_S(\bm{X_T}) = \begin{cases} \min \bm{T} & \bm{T}_{C_S} = \varnothing \\
        \max \bm{T}_{C_S} & \text{else}    \end{cases} \\
        t_C & = T_C(\bm{X_T}) = \begin{cases} t_{\underline{C}}(\max \bm{T}) & \bm{T}_{C_C} = \varnothing \\
        \min \bm{T}_{C_C} & \text{else}    \end{cases} \\
        t_A & = T_A(\bm{X_T}) = \begin{cases} \max \bm{T} + t_{\epsilon} & \bm{T}_{C_A} = \varnothing \\
        \min \bm{T}_{C_A} & \text{else}    \end{cases} \\ \label{eq:char_time_extract}
    \end{aligned}\end{equation}
    Here, a sample is excluded from the dataset, if no decision can be observed (i.e., if $\bm{T}_{C_C} = \varnothing \, \land \, \bm{T}_{C_A} = \varnothing$).
    \item The binary decision $a$, with $a=1$ for accepted gaps ($t_A <t_C$) and $a=0$ for rejected gaps, is extracted.
    \item $t_{\text{crit}}$ is extracted next, with
    \begin{equation*}
        t_{\text{crit}} = \begin{cases}    
        t_S & \Delta t_D (t_S) \leq 0 \\
        t_A + t_{\epsilon} & \min \{ \Delta t_D (t) \, \vert \,t_S \leq t < t_A \} > 0 \\
        t_D   & \text{else}
        \end{cases}\, , 
    \end{equation*}
    where
    \begin{equation*}
        t_D = \min \left\{ t \, \vert \, t > t_S \land  \Delta t_D (t) = 0 \right\}
    \end{equation*}
    satisfies both requirements in Equation \eqref{eq:crit_cond1}.
    \item The time of prediction $t_0$ is calculated accordingly to the method chosen previously (\ref{sec:t0_methods}). Only samples that meet the condition
    \begin{equation}
        t_S \leq t_0 < \min \left\{t_A, t_{\text{crit}} \right\} 
        \label{eq:t0_cond}
    \end{equation}
    are included in the final dataset, to ensure that gaps are already offered, $V_T$ has not made a decision yet, and that the prediction is still useful.
    \item The number of input time-steps $n_I$ and the time-step size $\delta t$ are chosen. 
    \item 
    One also has to determine the number of output time steps $n_O$:
    \begin{equation}
        n_O = \left \lceil {t_C - t_0 \over{\delta t}}   \right\rceil 
        \label{eq:NO}
    \end{equation}
    The resulting prediction horizon $n_O \delta t$ is therefore large enough to see the outcome of the gap acceptance scenario, i.e. the acceptance or closing of the gap.
    \item Based on $t_0$, $n_I$, $n_O$, and $\delta t$, the time-steps for input and output data are selected, named $\bm{T}_I$ and $\bm{T}_O$ respectively:
    \begin{equation}\begin{aligned}
        \bm{T}_I & = \left\{ t_0 + i \delta t \,\vert\,  i \in \{-n_I + 1, \hdots, 0    \right\} \\
        \bm{T}_O & = \left\{ t_0 + i \delta t \,\vert\,  i \in \{ 1, \hdots, n_O\} \right\}  \label{eq:TI_TO}
    \end{aligned}\end{equation}
    \item For those time-steps, the input trajectories $\bm{X}_{\bm{T}_I}$ and output trajectories $\bm{X}_{\bm{T}_O}$ are extracted from $\bm{X_T}$, using interpolation if necessary. 
    \item Certain domain information $k$ is collected, for instance, the location at which the trajectories $\bm{X_T}$ were collected or the test subjects involved in gathering the data.
\end{enumerate}
The input data $\bm{D_I}$ then includes from each sample the input trajectory $\bm{X}_{\bm{T}_I}$ and the corresponding time-steps $\bm{T}_I$. Meanwhile, the output data $\bm{D_O}$ takes the output trajectory $\bm{X}_{\bm{T}_O}$ and the corresponding time-steps $\bm{T}_O$, as well as the binary decision $a$, the time of accepting the gap $t_A$, and the domain information $k$ from each sample. 

\subsection{Creating training and testing set --- Splitting method $S$} \label{sec:SM}
To fulfill requirement \hyperref[req:2]{\textbf{R 2}}, the splitting method $S$, separating the given samples created in the previous step (\ref{sec:DI}) into training and testing sets, is a crucial part of the framework. As this functionality should be independent of the scenario, it is part of the separate splitting module.
Examples for this range from random splitting to methods taking into account all the information in $\bm{D_I}$ and $\bm{D_O}$.

Besides the potential similarity measure $\zeta$ of training samples to the training set, this functionality creates the training data $\bm{D}_{\bm{I},\text{train}}$ and $\bm{D}_{\bm{O},\text{train}} $ as well as the test data $\bm{D}_{\bm{I},\text{test}}$ and $\bm{D}_{\bm{O},\text{test}}$. 

\subsection{Training the model on the training set --- Model $M$} \label{sec:MT}
After splitting the samples into training and testing sets (\ref{sec:SM}), the model has to be trained on the training set ($\bm{D}_{I,\text{train}}$, $\bm{D}_{O,\text{train}}$), which is one of the functionalities of the model module that has to be individually implemented for each model. If the model, for instance, requires input velocities, extracting those from the given position data $\bm{X}_{\bm{T}_I}$ and $\bm{X}_{\bm{T}_O}$ is done here. 

\subsection{Making the predictions for the testing set --- Model $M$} \label{sec:MP}
For every sample from the input testing set $\bm{D}_{\bm{I},\text{test}}$, a prediction $\bm{d}_{\text{pred}}$ is made by the model trained previously (\ref{sec:MT}), with all $\bm{d}_{\text{pred}}$ constituting the set of predictions $\bm{D}_{\bm{OP},\text{test}}$. As such predictions rely on a trained model, this functionality is also part of the model module. Depending on this model, each prediction $\bm{d}_{\text{pred}}$ might take different forms. Three different forms of stochastic predictions are implemented into the framework.
\begin{itemize}
    \item Binary prediction: $\bm{d}_{\text{pred}} = a_\text{pred}$, i.e., only the probability $a_\text{pred} \in [0,1]$ of $V_T$ accepting the offered gap is predicted.
    \item Timing prediction: $\bm{d}_{\text{pred}} = \{a_\text{pred}, \bm{t}_{A,\text{pred}}\}$, i.e., not only $a_\text{pred}$ is predicted but also the time $\bm{t}_{A,\text{pred}}$ at which the gap acceptance might take place (Appendix~\ref{App:details_MP}). 
    \item Trajectory prediction: $\bm{d}_{\text{pred}} = \bm{X}_{\bm{T}_O,\text{pred}}$, i.e., the full trajectory of $V_T$ is predicted. This prediction consists of $n_p$ trajectories $\bm{X}_{\bm{T}_O,p}$---all equally likely---to represent probabilistic outputs: 
    \begin{equation*}
        \bm{X}_{\bm{T}_O, \text{pred} } = \left\{ \bm{X}_{\bm{T}_O,1}, \hdots, \bm{X}_{\bm{T}_O,n_p} \right\} 
    \end{equation*} 
\end{itemize}

\subsection{Transforming the predictions --- Dataset $D$} \label{sec:DT}
\begin{table*}[ht!]
    \caption{Transformation between the three possible prediction types $\bm{d}_{\text{pred}}$ implemented into the proposed framework, using the three implemented functions $T_i$.} 
    \centering
    \begin{tabularx}{0.95 \textwidth}{X |Y Y Y}
    \toprule
    Input $\Rightarrow$ Output    & $a_\text{pred}$ & $\{a_\text{pred}, \bm{t}_{A,\text{pred}}\}$ & $\bm{X}_{\bm{T}_O, \text{pred}}$\\ \midrule
    Binary prediction $a_\text{pred}$ & --- & $\left\{a_\text{pred}, T_3 \left(a_\text{pred} \right)\right\}$ & $T_2 \left(\left\{a_\text{pred}, T_3 \left(a_\text{pred} \right)\right\} \right)$ \\ 
    Timing prediction $\{a_\text{pred}, \bm{t}_{A,\text{pred}}\}$ & --- &  --- & $T_2\left(\left\{a_\text{pred}, \bm{t}_{A,\text{pred}}\right\} \right)$  \\ 
    Trajectory prediction $\bm{X}_{\bm{T}_O,\text{pred}}$ &   in $T_1 \left(\bm{X}_{\bm{T}_O,\text{pred}} \right)$ &  $T_1 \left(\bm{X}_{\bm{T}_O,\text{pred}} \right)$ & ---\\ 
    \bottomrule
    \end{tabularx}
    \label{tab:transform}
\end{table*}

To fulfill requirement \hyperref[req:3]{\textbf{R 3}}, we then must be able to transform a prediction $\bm{d}_{\text{pred}}$ from the previous step (\ref{sec:MP}) to another prediction form if necessary. This functionality is a part of a specific dataset, as this requires the context information to, for example, classify different trajectories as accepted or rejected gap.

To facilitate those transformations, three different functions $T_i$ (exact implementation in Appendix~\ref{App:details_DT}) are needed, as can be seen in Table~\ref{tab:transform}:
\begin{itemize}[align = left, leftmargin=1cm, itemindent = 0cm, labelwidth = 0.9cm, labelsep = 0.1cm]
    \item[$T_1$:] takes the trajectory prediction $\bm{X}_{\bm{T}_O,\text{pred}}$ and then provides $\{a_\text{pred}, \bm{t}_{A,\text{pred}}\}$. This is similar to the extraction of the time points in Section~\ref{sec:DI}.
    \item[$T_2$:] takes the prediction $\{a_\text{pred}, \bm{t}_{A,\text{pred}}\}$ and then provides the trajectory prediction $\bm{X}_{\bm{T}_O,\text{pred}}$, consisting of $n_p$ trajectories from the predictions of two conditional trajectory prediction models trained only on accepted and rejected gaps respectively. These are selected so that $T_1(\bm{X}_{\bm{T}_O,\text{pred}})$ results in the original inputs. 
    \item[$T_3$:] takes a binary prediction $a_\text{pred}$ and provides the predicted time of accepting the gap $\bm{t}_{A,\text{pred}}$, by extracting it from the prediction of a trajectory prediction model trained only on accepted gaps.
\end{itemize}

\subsection{Evaluating the predictions --- Metric $E$}\label{sec:EV}
This functionality---the main part of the metric module---implements the performance evaluation, comparing the actual outputs $\bm{D}_{\bm{O},\text{test}}$ (from \ref{sec:SM}) with the predicted outputs $\bm{D}_{\bm{OP},\text{test}}$ (\ref{sec:DT}). It returns either a combined value $F$ or instead a separate value for each sample $\bm{d}_{\text{pred}} \in \bm{D}_{\bm{OP},\text{test}}$, resulting in the output $\bm{F}$.

\section{Benchmark implementation}
\label{sec:Used_modules}
We implemented the framework described above by linking together several datasets, models, splitting methods, and metrics. These were chosen not to comprehensively cover all possible gap acceptance scenarios and prediction models but to demonstrate the flexibility and utility of the proposed framework. Still, our implementation can already serve as a benchmark for new prediction models. This section only presents an overview of the implementation, with full technical specification provided in \href{https://github.com/julianschumann/Framework-for-benchmarking-gap-acceptance/blob/main/Framework/Benchmark-Implementation.pdf}{supplementary materials}.

\subsection{Datasets}
\begin{table*}[ht!]
    \caption{The number of accepted gaps $N_A$ and rejected gaps $N_{\neg A}$ in the implemented datasets, in the form: $N_A - N_{\neg A}$ (Median $t_{\underline{C}}\left(t_0\right)$). The numbers depend on the method for choosing the time of prediction $t_0$} 
    \centering
    \begin{tabularx}{0.95\textwidth}{X |Y Y Y}
    
    \toprule
    
    Dataset    & {Initial gaps at their opening} & {Gaps with fixed size} & {Critical gaps} 
    \\ \midrule
    highD (Lane changes) & $1406 - 7026$ ($\SI{8.9}{s}$)\textcolor{white}{0} & $\textcolor{white}{0}461 - 1568$ ($\SI{11.8}{s}$) & $\textcolor{white}{000}0 - 7025$ ($\SI{0.8}{s}$)\textcolor{white}{0}
    \\ 
    highD (Lane changes - restricted) & $1406 - 1001$ ($\SI{12.9}{s}$) & $\textcolor{white}{0}392 - 241\textcolor{white}{0}$ ($\SI{8.7}{s}$)\textcolor{white}{0} & $\textcolor{white}{000}0 - 1000$ ($\SI{0.7}{s}$)\textcolor{white}{0}
    \\
    rounD (Roundabout) & $\textcolor{white}{0}662 - 917\textcolor{white}{0}$ ($\SI{2.5}{s}$)\textcolor{white}{0} & $\textcolor{white}{0}168 - 168\textcolor{white}{0}$ ($\SI{2.9}{s}$)\textcolor{white}{0} & $\textcolor{white}{00}33 - 913\textcolor{white}{0}$ ($\SI{1.0}{s}$)\textcolor{white}{0}
    \\ 
    L-GAP (Left turns) & $\textcolor{white}{0}703 - 724\textcolor{white}{0}$ ($\SI{4.6}{s}$)\textcolor{white}{0} & $\textcolor{white}{0}496 - 572\textcolor{white}{0}$ ($\SI{3.5}{s}$)\textcolor{white}{0} & $\textcolor{white}{0}369 - 723\textcolor{white}{0}$ ($\SI{2.3}{s}$)\textcolor{white}{0}
    \\ \bottomrule
    \end{tabularx}
    \label{tab:data_size}
\end{table*}
Different datasets are implemented into the framework (Table~\ref{tab:data_size}), including data recorded on real roads as well as data from a driving simulator study. The naturalistic datasets used here are captured by drones and distinguished by accurate position labeling. They cover lane changes on German highways (the \emph{highD} dataset~\cite{krajewski_highd_2018}) and roundabouts (the \emph{rounD} dataset~\cite{krajewski_round_2020}). The L-GAP dataset covers left turns at unsignalized intersections through oncoming traffic recorded in a driving simulator~\cite{zgonnikov_should_2022}. It has been chosen due to the simplicity of its environment, contrasting the more complex scenarios in the naturalistic datasets.

\subsubsection{Lane changes}
\label{sec:HLC}
Here we focus on lane changes of the target vehicle $V_T$ toward a faster lane to the left, along which the ego vehicle $V_E$ driving there has the right of way. While it could be argued that predictions in such situations could be simply based on turn signals, one cannot rely on human drivers to correctly use these~\cite{yang_examining_2019}. As a source of lane change data, we used two versions of the \emph{highD} dataset, \textit{full} and \textit{restricted}.

The full \emph{highD} dataset, not employing any filters, is heavily biased toward trajectories without a lane change. This is not a problem per se, but in such trajectories it is not known whether the target vehicle $V_T$ even had an intention to change lanes (i.e., if there was a gap acceptance situation in the first place). For this reason, in addition to the full \emph{highD} dataset, we added a restricted version of it which only included samples for which it can be inferred that the target vehicle $V_T$ indeed considered a lane change. Criteria are either a lane change of $V_T$ after $V_E$ has passed or $V_T$ braking to not collide with the preceding vehicle instead of changing lanes. 
Still, in both versions of \emph{highD}, the gaps are always accepted with large safety margins (Table~\ref{tab:data_size}).

\subsubsection{Roundabout}
In the \emph{rounD} dataset, the target vehicle $V_T$ has to enter a roundabout, which it can do in front of or behind the ego vehicle $V_E$ already in the roundabout. As the trajectories are recorded in Germany, the ego vehicle inside the roundabout has the right of way.
Compared to \emph{highD}, this dataset is far more balanced between accepted and rejected gaps, but still only includes few critically accepted gaps.

\subsubsection{Left turns} 
In the \emph{L-GAP} dataset~\cite{zgonnikov_should_2022}, the driver of the target vehicle $V_T$ intends to turn left at an intersection. The driver had to decide whether to do this in front of or behind the ego vehicle $V_E$ approaching the intersection from the opposite direction with the right of way. While the number of samples in this dataset is comparatively small, they are relatively balanced between accepted and rejected gaps. Also, they include many gaps accepted after $t_{\text{crit}}$ (Table~\ref{tab:data_size}).
Nonetheless, as $V_T$ starts in an idling position at some distance to the contested area, this might not be the most challenging dataset, as an onset of movement before $t_0$ in most cases is an apparent indicator of $V_T$ intending to accept the gap.

\subsection{Test-train Splitting Methods}
Two splitting methods are implemented, without a method for calculating the similarity measure $\bm{\zeta}$. Nonetheless, to enable at least a qualitative approximation of a model's robustness, the methods are designed to produce testing sets of varying difficulty for the prediction models.

The easier variant performs a stratified random splitting, while the second, more extreme method sorts the most unintuitive behavior of the target vehicle into the testing set (e.g. accepting a very small gap or rejecting a very large gap).

In both cases, the testing set includes $20\%$ of the samples and the training set the remaining $80\%$.

\subsection{Models}
The benchmark includes two state-of-the-art trajectory prediction models
\begin{itemize}
    \item \emph{Trajectron++} (also referred to as \emph{T+}), a deep-learning model mainly based on long-short-term memory cells~\cite{salzmann_trajectron_2020}.
    \item \emph{AgentFormer} (\emph{AF}), a deep-learning model based on transformers~\cite{yuan_agentformer_2021}. Compared to \emph{T+}, it has ten times more trainable parameters.
\end{itemize}

For the binary prediction models for gap acceptance, there is, as mentioned above, a lack of a common benchmark, making the models' selection more contentious. Four models have been selected nonetheless:
\begin{itemize}
    \item Logistic regression (\emph{LR}) is commonly used for predicting human gap acceptance decisions~\cite{theofilatos_cross_2021} and is therefore included as a simple baseline.
    \item Random forests (\emph{RF}) have been shown to outperform other approaches such as logistic regression and standard decision trees in gap acceptance prediction~\cite{mafi_analysis_2018}.
    \item Deep belief networks (\emph{DB}), also used previously to predict human gap acceptance decisions~\cite{xie_data-driven_2019}. 
    \item A metaheuristic model based on combining all other five models above (\emph{MH}); previously a similar approach for lane changes has been shown to outperform each of the models included in it~\cite{khelfa_predicting_2023}.
\end{itemize}
The benchmark does not include any dedicated timing prediction models yet, as their primary representative, the drift-diffusion model~\cite{zgonnikov_should_2022, pekkanen_variable-drift_2022}, can currently not be trained on datasets with a large number of unique samples in a reasonable amount of time. Nonetheless, to allow for future expansion of the benchmark, the framework has been designed with such models in mind.

\subsection{Evaluation Metrics}
\label{sec:used_metrics}
We have included several metrics that characterize models in terms of the quality of binary predictions (accept/reject gap) as well as full trajectory predictions. The following metrics are commonly used in the literature:
\begin{itemize}
    \item Accuracy: This metric is a widespread method to evaluate the performance for binary prediction models~\cite{mafi_analysis_2018,xie_data-driven_2019,khelfa_predicting_2023}. However, accuracy is a symmetric metric, i.e. it is unable to differentiate between false negative and false positive predictions. Consequently, it is best used in cases where $t_0 \ll t_{\text{crit}}$, as the consequences of false predictions are not too different there (Appendix \ref{App:details_EV}).
    \item \emph{AUC}: This metric for binary prediction models, the Area Under Curve of a receiver operating characteristics curve, addresses one central point of criticism of the accuracy metric, namely its sensitivity to biases in the testing set. Nonetheless, like the accuracy metric, it does not consider the potentially differing severity of false predictions. Hence, we only apply it to rate a prediction model's performance when $t_0 \ll t_{\text{crit}}$.
    \item $\text{\emph{ADE}}_{\beta}$ and $\text{\emph{FDE}}_{\beta}$: These metrics, the Average or Final Displacement Error of the $n_p \beta$ least erroneous predicted trajectories, are commonly applied to trajectory predictions~\cite{salzmann_trajectron_2020,yuan_agentformer_2021, giuliari_transformer_2021}, with $\beta=1$ (i.e., average for all predictions) or $\beta = 0.05$ (mimicking the ``best-of-20'' metric used in~\cite{salzmann_trajectron_2020}) being used in this work. As these metrics also does not take into account the severity of different false predictions~\cite{ivanovic_injecting_2022, farid_task-relevant_2022} and requires equally long prediction horizons as well, they are only applied for constant gap sizes ($t_0 = \min \{t \,\vert \, t_{\underline{C}}(t) - t = \Delta t\} \ll t_{\text{crit}}$). However, due to their similarity, only the \emph{ADE} is discussed furhter in this work, while the result for \emph{FDE} can instead be found in the \href{https://github.com/julianschumann/Framework-for-benchmarking-gap-acceptance/blob/main/Framework/Benchmark-Implementation.pdf}{supplementary materials}.
\end{itemize}

Similarly to the \emph{FDE}, an additional metric, namely the miss rate (\emph{MR}) is provided in the supplementary materials as well. Furthermore, we propose a novel metric that considers the potential consequences of a wrong prediction.
\begin{itemize}
    \item \emph{TNR-PR}: The True Negative Rate under Perfect Recall is a metric applied to binary predictions, with the explicit goal to consider the vastly different consequences of false negative and positive predictions at $t_0 \approx t_{\text{crit}}$ (i.e., a potential accident vs unnecessary braking, see Appendix~\ref{App:details_EV}). To that end, the threshold for classifying a prediction $a_{\text{pred}}$ as positive is set as low as necessary to achieve perfect recall on the test set (i.e., there are no false negative predictions). It then estimates the usefulness of prediction models in improving the efficiency of path planning by evaluating their true negative rate (TNR) under this decision threshold. This metric is therefore equivalent to the likelihood that a prediction model can prevent needless braking while guaranteeing safe interactions. As earlier predictions do not necessitate a model with perfect recall, due to available time to wait for further data, this metric is only applied to predictions made at the last possible time (i.e., $t_0 = t_{\text{crit}} - t_{\epsilon}$).
\end{itemize}

\section{Results}
\begin{figure*}[!ht]
\centering
\includegraphics[]{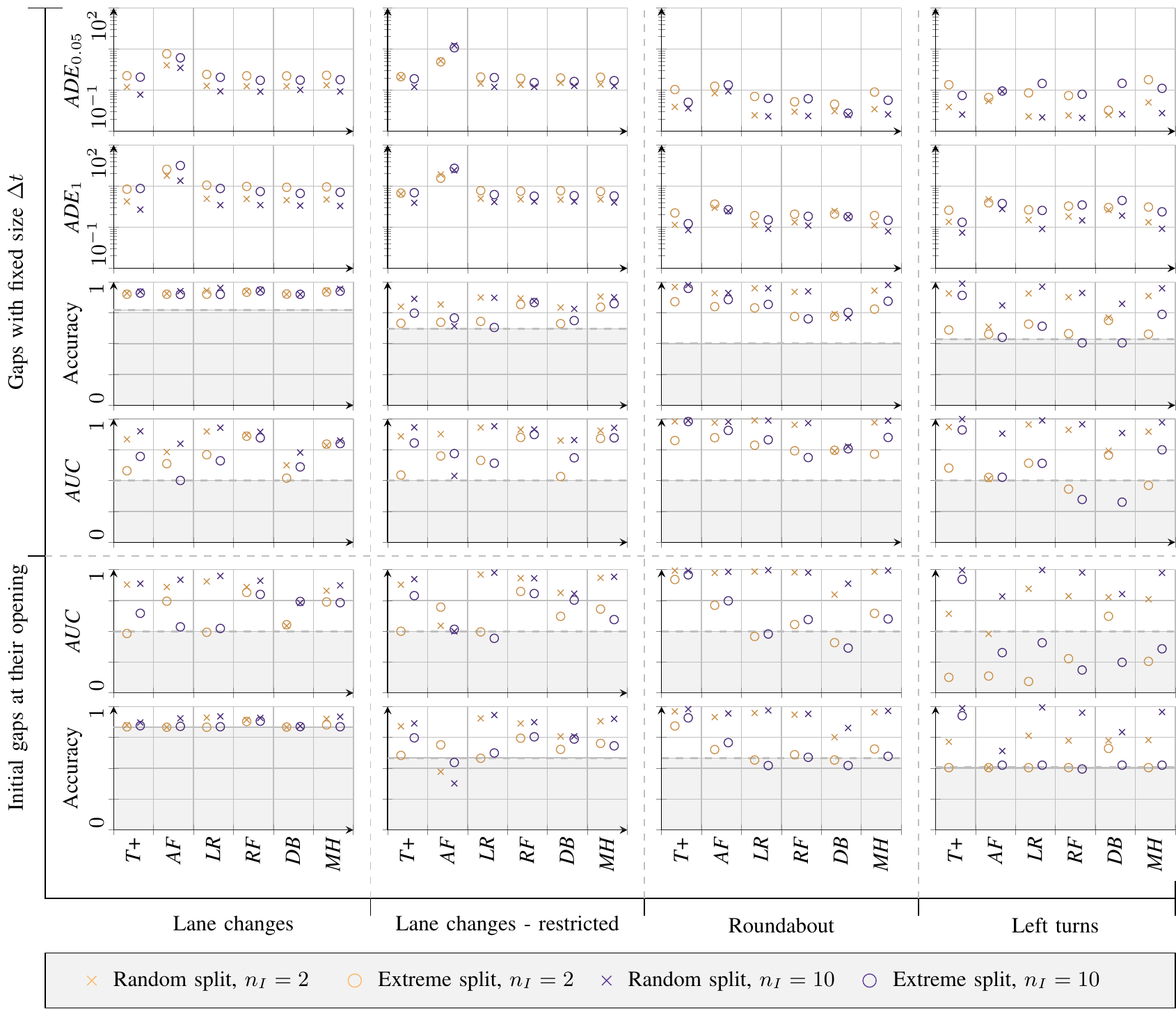}
\caption{The results of evaluating behavior prediction models in gap acceptance scenarios on different datasets, with prediction being made either at the initial opening of the gap ($t_0 = t_S$) or with fixed gap sizes ($t_0 = \min \{t \,\vert \, t_{\underline{C}}(t) - t = \Delta t\}$). The color indicates the number of input time-steps $n_I$ given to the models, and the marker type denotes the splitting method, which can be \textit{random} or \textit{extreme}. The dashed gray lines indicate the performance $F_r$ of a uniformly random binary predictor. All results can be found in the form of tables in the \href{https://github.com/julianschumann/Framework-for-benchmarking-gap-acceptance/blob/main/Framework/Benchmark-Implementation.pdf}{supplementary materials}.}
\label{fig:main_results}
\end{figure*}
\begin{figure}[!ht]
\centering
\includegraphics[]{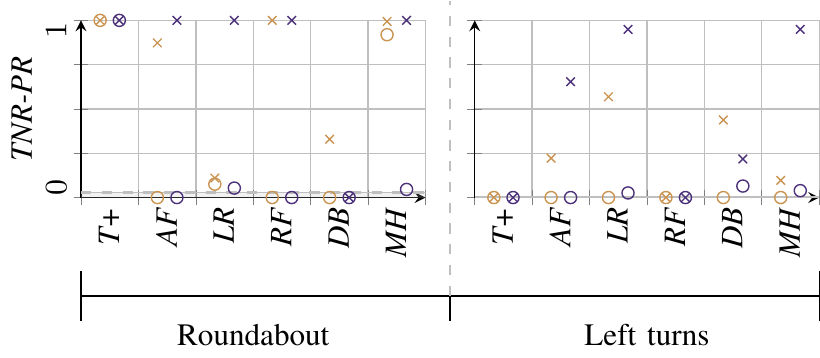}
\caption{The true negative rate under perfect recall (\emph{TNR-PR}) of different prediction models, using the same visualizations as in Figure~\ref{fig:main_results}, tested on the last useful predictions ($t_0 = t_{\text{crit}} - t_{\epsilon}$).}
\label{fig:TNR-PR_results}
\end{figure}

Our benchmark provides insights into performance of tested models under different conditions (Figures~\ref{fig:main_results}, \ref{fig:TNR-PR_results}). In this section, we first discuss our findings regarding the prediction performance across models depending on specific prediction problems, which are defined by number of input time steps $n_I$, the splitting method, the scenarios, and the prediction time $t_0$. We will then discuss the differences between individual models. 

\subsection{Prediction problem's influence on model performance}
As one may expect, when provided with more input time steps ($n_I = 10$ vs. $n_I = 2$), i.e., more information is provided to extract signs of future behavior from, the models' predictions were generally better. Second, performance of the models tested on the most unintuitive samples (the extreme splitting case) was worse than in the random splitting case. There, the models have to extrapolate to situations outside the training domain, a typically far more difficult task than the interpolation inside the training domain needed for prediction on random test samples~\cite{barnard_extrapolation_1992}. The poor performance on unintuitive samples is especially pronounced when looking at the \emph{TNR-PR} at critical gaps (Figure~\ref{fig:TNR-PR_results}), where no model could substantially outperform a random predictor on both datasets.

When comparing the predictions of models on the two lane-change datasets (the two left columns in Figure \ref{fig:main_results}), it can be expected that the models' performance should be better on the restricted one, due to the removal of many large gaps that were unintuitively rejected. However, this was only observed in the \emph{AUC} metric, while the opposite could be observed for Accuracy (caused by the exploitation of the larger bias towards rejected gaps in the unrestricted lane change dataset, an explanation supported by the high \emph{MR} (\href{https://github.com/julianschumann/Framework-for-benchmarking-gap-acceptance/blob/main/Framework/Benchmark-Implementation.pdf}{supplementary materials}) in those cases) and \emph{ADE}. Therefore, it can be assumed that \emph{AUC} is the most reliable metric here, a result which is supported for example by Huang and Ling~\cite{huang_using_2005}; this metric will be our main focus from here on.

When comparing the difficulty of different scenarios, it can be seen that, generally, the prediction of human behavior at roundabouts seems to be the easiest, having the best \emph{AUC} values in 30 of 48 cases (each two prediction times, input steps, and splitting methods on six models). A possible explanation here might be the short prediction horizon of less than three seconds (Table \ref{tab:data_size}), which leaves less room for the target vehicle to behave unexpectedly, although scenario-specific reasons cannot be excluded. This explanation is also supported by the finding that---at least for $n_I = 10$---the scenario with the next shortest gap, the left turns, gets the best results out of the remaining three datasets in 12 of 24 cases. This is not the case for $n_I = 2$, where the left-turn dataset is the worst of all four datasets in 15 of 24 cases. This can be explained by the fact that in this scenario, the target vehicle starts in an idling position, from which not much information can be gained. Technically, predictions from both $n_I = 2$ and $n_I = 10$ should be made at the same time, but due to the lack of trajectory data before $t_S$ in this dataset, the predictions are made in fact at $t_0 > t_S + (n_I - 1) * \delta t$. Therefore, prediction for $n_I = 10$ are generally made later, which leaves more room for the onset of motion, making predictions easier.

When comparing the \emph{AUC} at different prediction times $t_0$, it can be observed that on those datasets with a lower median $t_{\underline{C}}(t_0)$ (Table \ref{tab:data_size}) the prediction were better more often than not, in 59 of 96 cases. This again lends credence to the hypothesis that earlier predictions with longer prediction horizons are more challenging. In the case of trajectory prediction models, this seems logical, as a longer prediction horizon leaves more time for prediction errors to propagate and compound on each other. There is also the higher probability that the duration between $t_0$ and the actual human decision is larger, which leads to a higher probability that there are no indications of that decision in the human behavior yet.

\subsection{Differences between models}
Comparing the trajectory prediction models, we found that the Trajectron++ model (\emph{T+}) consistently outperforms the AgentFormer model (\emph{AF}). This is surprising, as \emph{AF} previously outperformed \emph{T+} on pedestrian trajectory prediction benchmarks~\cite{yuan_agentformer_2021}. This contradiction might be explained by over-fitting the many trainable parameters for \emph{AF} on relatively small datasets here.
Meanwhile, the logistic regression ($LR$) model is often the most promising approach for binary prediction models (best \emph{AUC} value in 18 of 32 cases, and best \emph{TNR-PR} value in 5 out of 8 cases), especially when tested on random samples, where it has the best \emph{AUC} in all possible sixteen cases. Together, those results indicate that increasing the complexity of such models and their number of parameters might not be a panacea, with simpler models being more promising, especially if datasets are relatively small.

When we compare binary models against trajectory prediction models, we can observe differing behavior for different scenarios. On the one hand, the best performance on the lane-change datasets is generally achieved by a binary prediction model (the best \emph{AUC} values come from binary models in all 16 cases), while on the other hand, similar performance can be observed on the other two datasets. One main difference here is the prediction horizon ($\Delta t \approx \SI{10}{s}$ and $\Delta t < \SI{4}{s}$ respectively, see Table \ref{tab:data_size}), which might indicate that trajectory prediction models are more impaired by such longer prediction horizon than binary models. That the average displacement errors are much more noticeable in the lane change scenario also supports this explanation, further showing the difficulties of using trajectory prediction under those conditions. However, a deeper analysis into the causes for these observations when comparing models would likely require a number of ablation studies, which is outside the scope of this work.

Lastly, when evaluating the promise of including binary prediction models into trajectory prediction models (relying on the transformation function $T_2$ from Table Table~\ref{tab:transform}), we can hold that the benefits are mostly negligible if existent at all, except at roundabout and left turns, where some improvements can be seen in average displacement errors. Nonetheless, due to the problems with that metric, more than those results are needed to render a final judgment. However, due to the small size of datasets and the low number of models, the results discussed here should be treated with care.

\section{Conclusion}
We proposed a framework that connects previously disparate datasets, models, and metrics in the benchmark for testing behavior prediction models in gap acceptance scenarios. We demonstrated its potential and flexibility by comparing two state-of-the-art trajectory prediction models with several binary gap acceptance models. 
Additionally, we showed that relying on the characteristic time points of gap acceptance scenarios to select the most unintuitive samples in the splitting module is a promising approach to analyzing model generalization, as seen by the general decrease in performance for models trained on those samples.
Our framework is open-source and specifically designed in a modular way to simplify adding new datasets, splitting methods, models, and evaluation metrics, which allows researchers to easily expand it in future. This can speed up the testing of new models, only requiring the adaption of the models' implementation to the format of our framework, while previously one would need to write separate code for every model applied to every scenario. 

One particularly important addition to the benchmark would be datasets containing more critically accepted gaps, as this would allow for an increased meaningfulness of metrics applied to last useful predictions. Additionally, a metric better aligned with the main purpose of a prediction model as a part of an autonomous vehicle is still needed. Likewise, currently there exists no method for calculating the similarity between testing and training set $\zeta$; the future addition of this would permit a quantitative comparison of a model's robustness against unintuitive test samples. Lastly, one could expand the framework to provide scenario-independent inputs similar to $t_{\underline{C}}$, which would enable training a model on two unrelated datasets simultaneously, leading to a better estimation of the models generalizability.

We acknowledge that testing a model in gap acceptance scenarios alone is necessary, but not sufficient for justifying its usage in actual vehicles. Consequently, expanding the framework to non-gap-acceptance scenarios is an important avenue for future research. This will enable more holistic testing, although only for models predicting (and metrics evaluating) trajectories. Nonetheless, we argue that performance of models on non-gap acceptance scenarios should still be given lower priority compared to gap acceptance scenarios which are more safety-critical.

Our results resonate with the recent literature on hybrid AI~\cite{marcus_next_2020, van_bekkum_modular_2021}, showing that including binary prediction models in specific scenarios might make data-driven trajectory prediction models more reliable, especially in accurately predicting dangerous situations. However, especially for the unintuitive and safety-critical edge cases, most models often performed only slightly better than a random predictor at best. Therefore, there currently seems to be no model that a trajectory-planning algorithm can rely on to substantially increase the effectiveness of an autonomous vehicle's driving style in every scenario, necessitating further research into such models.

{\appendices
\renewcommand{\theequation}{A.\arabic{equation}}

\section{Detailing the framework for benchmarking gap acceptance models}
\label{App:framework_details}

\subsection{Influence of $t_0$ on the importance of predictions} \label{App:details_EV}
Due to the differing consequences of false negative and false positive predictions when using binary predictions $a_{\text{pred}}$, there are limitations on which metrics are usable at certain $t_0$. For $t_0 \ll t_{\text{crit}}$, the consequences of a wrong prediction are generally minor, as time is left to wait for future information before more significant changes to trajectory planning are necessary. Furthermore, even if the target vehicle would immediately accept the gap after $t_0$, the necessary response is likely neither uncomfortable nor risky. Consequently, symmetric metrics can be used here.

For $t_0 \approx t_{\text{crit}}$ however, no more time for further observations is left, resulting in far more severe consequences for both false positive and false negative predictions. A false positive prediction would unnecessarily result in a harsh and uncomfortable braking maneuver. Meanwhile, a wrong negative prediction leads to an unsafe gap acceptance maneuver, with the safety of the interaction between $V_E$ and $V_T$ no longer in the control of the autonomous vehicle $V_E$. Accidents or the need for dangerous emergency maneuvers, which could result in material damage or even bodily harm, are then possible. As the latter should be avoided at all costs, a false negative prediction at this time is far more consequential, which should be reflected in the evaluation metric. 

\subsection{Influence of $t_0$ on the size of the dataset}
\label{App:details_ET}
The method for determining the time $t_0$ can impact the size of the resulting dataset (Table~\ref{tab:data_size}) due to the condition from Equation~\eqref{eq:t0_cond}. Namely, for constant gap size ($t_0 = \min \{t \,\vert \, t_{\underline{C}}(t) - t = \Delta t\}$), the number of available samples will be reduced, as all gaps with an initial smaller gap size ($t_{\underline{C}}(t_S) - t_S < \Delta t$) will be excluded. The same is the case for gaps already accepted before $t_0$. For critical gap sizes instead (i.e., $t_0 = t_{\text{crit}} - t_{\epsilon}$), all gaps accepted before $t_{\text{crit}}$ will be excluded, leading to extremely biased datasets, sometimes even removing all accepted gaps.

\subsection{Detailing the timing prediction} \label{App:details_MP}
The predicted time $\bm{t}_{A,\text{pred}}$ is expressed using the decile values:
\begin{equation}
    \bm{t}_{A,\text{pred}} = \bm{Q}_9 (\mathcal{T}_A) = \left\{ Q_{\mathcal{T}_A}(p) \,\vert\, p \in \{0.1, 0.2, \hdots, 0.9\}\right\} \in \mathbb{R}^9 
    \label{eq:ta_pred}
\end{equation}
Here, $Q_{\mathcal{T}_A}$ is the quantile function associated with this underlying distribution of $\mathcal{T}_A$, which might for example be expressed as a set of individual time points $t_A$. This means that with a likelihood of $p$, $t_A < Q_{\mathcal{T}_A}(p)$ will be the case, if the human decides to accept the gap ($t_A < t_C$):
\begin{equation*}
    P\left(t_A < Q_{\mathcal{T}_A}(p) \, \vert \, t_A < t_C\right) = p
\end{equation*}
As the likelihood of accepting the gap is given by $a_{\text{pred}}$, one can get:
\begin{equation*}\begin{aligned}
    P\left(t_A < Q_{\mathcal{T}_A}(p)\right) & = P\left(t_A < Q_{\mathcal{T}_A}(p) \, \vert \, t_A < t_C\right) \, P\left(t_A < t_C\right) \\
    & = p \, a_{\text{pred}}
\end{aligned}\end{equation*}
As $t_A$ is predicted under the assumption that the gap is accepted, every decile value in $\bm{t}_{A,\text{pred}}$ should be smaller than $t_C$ and larger than $t_S$.

\subsection{Transforming the predictions} \label{App:details_DT}
When implementing the transformation of a prediction $\bm{d}_{\text{pred}}$ into another form, two instances of the trajectory prediction model \emph{Trajectron++}~\cite{salzmann_trajectron_2020} are used, namely $M_A$, trained on all samples from the specific dataset $D$ ($\bm{D_I}$ and $\bm{D_O}$) where $a = 1$, and $M_{\neg A}$, trained on the remaining samples where $a = 0$. Furthermore, the function $f_a$ is defined, which extracts the gap acceptance decision from a single predicted trajectory, using $T_A$ from equation \eqref{eq:char_time_extract} (assuming $t_C \approx \max \bm{T}_O$ based on Equations~\eqref{eq:NO} and~\eqref{eq:TI_TO}).
\begin{equation}
    f_a\left(\bm{X}_{\bm{T}_O,p}\right) = \begin{cases} 1 & T_A\left(\bm{X}_{\bm{T}_O,p}\right) < \max \bm{T}_O \\
    0 & \text{else} \end{cases}
\end{equation}
\begin{itemize}[align = left, leftmargin=1cm, itemindent = 0cm, labelwidth = 0.9cm, labelsep = 0.1cm]
    \item[$T_1$:] The use of $f_a$  and Equation~\eqref{eq:ta_pred} results in 
    \begin{equation}\begin{aligned}
        a_{\text{pred}} &= {1\over{n_p}} \sum\limits_{p} f_a\left(\bm{X}_{\bm{T}_O,p}\right) \\
        \bm{t}_{A,\text{pred}} &= Q_9\left(\left\{T_A\left(\bm{X}_{\bm{T}_O,p}\right) \, \vert \, f_a\left(\bm{X}_{\bm{T}_O,p}\right) = 1\right\} \right) \, .
    \end{aligned}\end{equation}
    
    \item[$T_2$:] $M_A$ and $M_{\neg A}$ are used to respectively predict two sets of trajectories, namely $\bm{X}_{\bm{T_O},\text{pred},A}$ and $\bm{X}_{\bm{T_O}, \text{pred},\neg A}$. The final set of trajectories $\bm{X}_{\bm{T_O},\text{pred}}$ is assembled from select trajectories out of these two sets:
    \begin{equation*}\begin{aligned}
        \bm{X}_{\bm{T_O},\text{pred}} = \, &\{\bm{X}_{\bm{T_O},p,\neg A} \vert p \in \bm{R}_{\neg A} \} \, \cap \\ &\{ \bm{X}_{\bm{T_O}, p,A} \vert p \in \bm{R}_{A}\} \, .
    \end{aligned}\end{equation*}
    This selection is based on the random selection function $R(m,\bm{M}, \bm{W})$, which randomly selects $m$ samples from a set $\bm{M}$, with the probabilities of  selection being proportional to the the weights $\bm{W}$. Only samples that actually represent the desired decision are viable (e.g., in samples from $\bm{X}_{\bm{T_O}, p,A}$, the gap must be accepted):
    \begin{equation*}\begin{aligned}
        \bm{R}_{\neg A}&=   R (n_p (1 - a_{\text{pred}}), \{ p \,\vert \, f_a(\bm{X}_{\bm{T_O}, p,\neg A}) = 0\}, \bm{1} ) \\
        \bm{R}_{A}&=  R(n_p a_{\text{pred}}, \{ p \,\vert \, f_a(\bm{X}_{\bm{T_O}, p,A}) = 1\}, \bm{W}_A ) \, .
    \end{aligned}\end{equation*}
    Here, $\bm{W}_A$ is chosen so that the distribution of $t_A$ described by the decile values $t_{A,\text{pred},i}$ of $\bm{t}_{A, \text{pred}}$ (Equation~\eqref{eq:ta_pred}) is maintained:
    \begin{equation*}
        \sum\limits_{\left\{p \vert 
        t_{A,\text{pred},i} < T_A\left(\bm{X}_{\bm{T}_O,p}\right) < t_{A,\text{pred},i + 1}  
        \right\}} w_{A,p} = 1  \;\;  \forall i 
    \end{equation*}
    This approach of using conditional trajectory prediction models is inspired by Xie et al.~\cite{xie_data-driven_2019} and Hu et al.~\cite{hu_causal-based_2022}.
    \item[$T_3$:] Here, one uses $M_A$ to get generate $\bm{X}_{\bm{T_O}\text{pred},A}$, based on which one can get 
    \begin{equation*} \begin{aligned}
        \bm{t}_{A,\text{pred}} = Q_9\left(\left\{T_A\left(\bm{X}_{\bm{T}_O,p, A}\right) \, \vert \, f_a\left(\bm{X}_{\bm{T}_O,p, A}\right) = 1\right\} \right) \, .
    \end{aligned}\end{equation*}
    
\end{itemize}
}

\bibliographystyle{jabbrv_ieeetr}
\bibliography{IEEEabrv,zotero_library}

\begin{IEEEbiography}[{\includegraphics[width=1in,height=1.25in,clip,keepaspectratio]{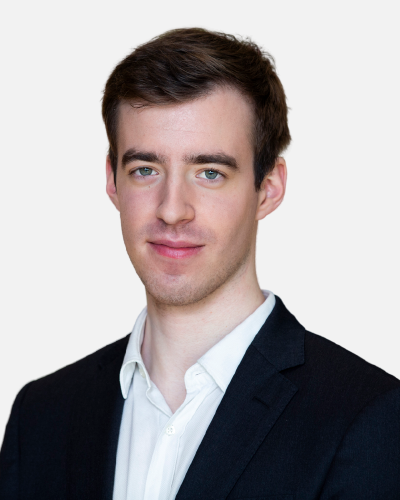}}]{Julian F. Schumann}
received the master's degree in 2021 in mechanical engineering from the TU Delft, The Netherlands, where since 2021, he is working toward the Ph.D. degree focusing on hybrid-AI models for behavior prediction of human traffic participants for use in automated vehicles. His research interests include modeling of human behavior, meaningful evaluation of such models, and the integration of knowledge-based and machine-learned models.
\end{IEEEbiography}
\begin{IEEEbiography}[{\includegraphics[width=1in,height=1.25in,clip,keepaspectratio]{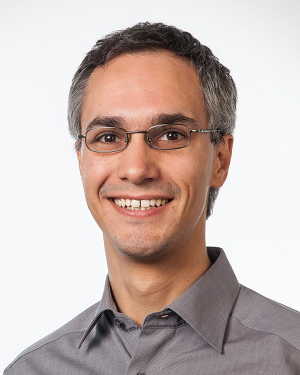}}]{Jens Kober}
is an associate professor at the TU Delft, Netherlands. He worked as a postdoctoral scholar jointly at the CoR-Lab, Bielefeld University, Germany and at the Honda Research Institute Europe, Germany. He graduated in 2012 with a PhD Degree in Engineering from TU Darmstadt. For his research he received the 2018 IEEE RAS Early Academic Career Award and the 2022 RSS Early Career Award. His research interests include motor skill learning, imitation learning, interactive learning, and machine learning for control.
\end{IEEEbiography}
\begin{IEEEbiography}[{\includegraphics[width=1in,height=1.25in,clip,keepaspectratio]{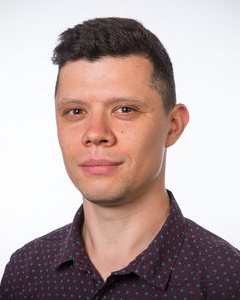}}]{Arkady Zgonnikov} received his Ph.D. degree from the University of Aizu, Japan, in 2014. Since then he worked as a postdoctoral researcher at the University of Galway, Ireland (funded by Irish Research Council) and Delft University of Technology, Netherlands (funded by the AiTech initiative). Since 2020, he has been an assistant professor at Delft University of Technology. His research interests include responsible AI and cognitive modeling, as well as applications of those to human-robot interactions in traffic and beyond. 
\end{IEEEbiography}

\end{document}